\pdfoutput=1
\documentclass[11pt]{article}

\usepackage[]{acl}

\usepackage[multiple]{footmisc}

\usepackage{times}
\usepackage{latexsym}

\usepackage{caption}
\usepackage{subcaption}

\usepackage{graphicx}

\usepackage[T1]{fontenc}
\usepackage[utf8]{inputenc}

\usepackage{microtype}

\title{KUCST@LT-EDI-ACL2022:\\Detecting Signs of Depression from Social Media Text}

\author{Manex Agirrezabal \and Janek Amann \\
  Centre for Language Technology \\
  Department of Nordic Studies and Linguistics \\
  University of Copenhagen \\
  \texttt{manex.aguirrezabal@hum.ku.dk}, \texttt{ja@developdiverse.com}
}

\begin{document}
\maketitle
\begin{abstract}
In this paper we present our approach for detecting signs of depression from social media text. Our model relies on word unigrams, part-of-speech tags, readabilitiy measures and the use of first, second or third person and the number of words. Our best model obtained a macro F1-score of $0.439$ and ranked 25th, out of 31 teams. We further take advantage of the interpretability of the Logistic Regression model and we make an attempt to interpret the model coefficients with the hope that these will be useful for further research on the topic.
\end{abstract}

\section{Introduction}

%Human beings show variations in their mood, but depression is a condition that affects further than just the mood. Depression is a common mental disorder and it is estimated that 5\% of adults are affected by it. Human beings have varying mood, but depression is a condition that affects further than just the mood. It can vary in degrees of intensity of its symptoms and when it is severe, it may become a serious health condition.

%The medical dictionary of Merriam Webster establishes that depression is a "\textit{mood disorder marked by varying degrees of sadness, despair, and loneliness that is typically accompanied by inactivity, guilt, loss of concentration, social withdrawal, sleep disturbances, and sometimes suicidal tendencies}".\footnote{\url{https://www.merriam-webster.com/dictionary/depression}}

Depression\footnote{\url{https://www.who.int/news-room/fact-sheets/detail/depression}}\footnote{\url{http://purl.bioontology.org/ontology/SNOMEDCT/35489007}} is a mental illness that affects to the 5\% of adults. The \textit{World Health Organization} states that depression is the leading cause of disability worldwide. Human beings have varying mood, but depression is a condition that affects further than solely the mood. Depending on the degree of intensity of its symptoms, it may become a serious health condition. In spite of the magnitude and risk, there is effective ways of treating mild, moderate and severe depression.

%Depression is a mood disorder that affects ... From Merriam Webster medical dictionary:\footnote{\url{https://www.merriam-webster.com/dictionary/depression}} a mood disorder marked by varying degrees of sadness, despair, and loneliness that is typically accompanied by inactivity, guilt, loss of concentration, social withdrawal, sleep disturbances, and sometimes suicidal tendencies 
%Further, from\footnote{\url{https://www.merriam-webster.com/dictionary/clinical\%20depression}}, a serious mood disorder involving one or more episodes of intense psychological depression or loss of interest or pleasure that lasts two or more weeks and is accompanied by irritability, fatigue, poor concentration, sleep disturbances, weight gain or loss, feelings of worthlessness or guilt, and sometimes suicidal tendencies : major depression, major depressive disorder
%Mild depression is a normal condition, a sadness brought on by life's circumstances. Clinical depression, on the other hand, is a sadness so deep that it can lead to suicide if left untreated. (Attributed to Brenda Lane Richardson by Merriam Webster)

In this paper we present our attempt for automatically classifying whether a social media post shows signs of moderate or severe depression. This is part of the \textit{Shared Task on Detecting Signs of Depression from Social Media Text} at the LT-EDI-2022 workshop \cite{depression2022st}. All our code is available in the following repository.\footnote{\url{https://github.com/manexagirrezabal/depression_detection_EDI2022}}

The paper is structured as follows. First we introduce some related work on the topic. Then, we introduce the data that we employed. We continue with the used features and the actual models that we trained. After that we present the results and briefly discuss the model coefficients, and finally we conclude the paper with some possible future directions.

\section{Related work}
There have been several attempts to model the language of people with depression. In some works the focus is on detection of social media posts from users with different degrees of depression and in some other cases, the goal was to analyze the language style of people with depression.

%In the work by \cite{coppersmith-etal-2015-clpsych}, the authors attempted to classify  users as (1) depression versus control, (2) PTSD versus control, and (3) depression versus PTSD. They worked on data from Twitter.

There is a large number of works that have attempted to detect depression from Social Media text. Some works employ Twitter \cite{coppersmith-etal-2015-clpsych,Choudhury_Gamon_Counts_Horvitz_2021,cavazos2016depression,mowery-etal-2016-towards,pirina-coltekin-2018-identifying,tadesse2019depression} and they work with different degrees of granularity with depression or depression-related symptoms.

%Twitter
%Similar task, but data has a different style. It is on Twitter, and we work on reddit data. They mostly used word ngrams. \cite{coppersmith-etal-2015-clpsych}
%https://aclanthology.org/W15-1204.pdf
%These guys also work on Twitter \cite{Choudhury_Gamon_Counts_Horvitz_2021}
%https://ojs.aaai.org/index.php/ICWSM/article/view/14432

%Analysis of tweets related to depression \cite{cavazos2016depression} 
%https://www.sciencedirect.com/science/article/pii/S0747563215300996
%We cannot say, though, that they analyzed the language of depressed people, as they did not distinguish between depressed or non depressed users. They performed an interesting analysis of tweets that talk about depression.

%In this work they did classification \cite{mowery-etal-2016-towards}
%https://aclanthology.org/W16-4320.pdf
%no evidence of depression or evidence of depression
%If there was evidence of depression, a depressive symptom
%If so, which was the symptom?
%depressed mood, disturbed sleep, or fatigue or loss of energy.
%They worked on Twitter, where the posts are in general shorter than the ones from reddit (max. 140 characters).

%A dataset from social media about depression \cite{losada2016depression} %\url{https://link.springer.com/chapter/10.1007/978-3-319-44564-9_3}

Other researchers have employed other social media that contain longer essays, such as Reddit \cite{ireland2020depression,iavarone2021depressionsuicide} for the detection of depression, by employing posts of users that were self-reported to have depression. Reddit posts have been further employed for, for instance, Bipolar disorder detection \cite{sekulic-etal-2018-just} or anxiety detection \cite{shen-rudzicz-2017-detecting}.

With regards to features, many works make use of the Linguistic Inquiry and Word Count (LIWC) \cite{pennebaker2007LinguisticIA}.
As in other Natural Language Processing related tasks, models based on contextual word embeddings have shown a good performance for depression detection, e.g. \cite{martinezcastano2020depression}. As they report, the performance of the model is high but the interpretability of the model could be improved.

Besides, the use of personal pronouns have been analyzed by many researchers. For instance in \newcite{rude2004languageuse}, they analyzed the language use of currently-depressed, formerly-depressed and never-depressed college students. Among other factors, they analyze the use of the first person pronoun ``I'' and they found that formerly-depressed and currently-depressed participants used the word ``I'' more often than the never-depressed participants. Furthermore, \newcite{Tackman2019DepressionNE} claim that depressive symptomatology is manifested in a greater use of the first-person singular pronoun and find a small but reliable positive correlation between depression and I-talk.

\section{Data}
\begin{figure}
    \centering
    \includegraphics[width=\columnwidth]{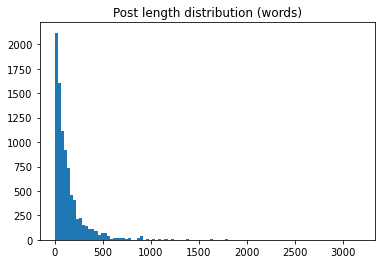}
    \caption{Histogram that shows the distribution of the length of social media posts in the current data set.}
    \label{fig:histogramwords}
\end{figure}

We make use of the data provided by the organizers of the shared task, built from Reddit posts \cite{kayalvizhi2022dataset}.\footnote{\url{https://competitions.codalab.org/competitions/36410}} Some of these posts have no depression signs, others show moderate depression signs and finally, there are the ones that show severe depression signs. The dataset contains $8891$ posts, from which the ones with no, moderate and severe depression signs are $1971$, $6019$ and $901$, respectively. All posts are written in English. Figure \ref{fig:histogramwords} shows a histogram with the length of the posts.

\section{Features and models}
In this section we present the features that we employed. Many of the features have been widely used for text classification and authorship analysis.

\paragraph{Words.} Bag of words as implemented by the CountVectorizer package from the \texttt{scikit-learn} library \cite{Pedregosa2011ScikitlearnML}. The expectation was that word usage might differ from depressed to non depressed users, and therefore, we expected that this feature would result beneficial.

\paragraph{Pos-tags.} We also included part-of-speech tags among the employed features. But, we did not incorporate them as single counts, but we normalized them in a way that we got a probability distribution of pos-tags. We simply counted the frequency of each pos-tag in each post and then normalized them using the \textit{softmax} function.

%POS tags. The distribution of pos tags. Why do you think they will make a difference in this?

\paragraph{Readability and style.} On top of that, we employ several readability and style related features as returned by a Python package called \texttt{readability}.\footnote{\url{https://github.com/andreasvc/readability}} This package includes readability metrics, such as the Automated Readability Index (ARI), Coleman-Liau, Dale-Chall, and so on, \footnote{Please refer to the Github repository for a full list of outcomes.} and some further stylistic features.

%Readability measures: We used the readability package available on Github\footnote{\url{https://github.com/andreasvc/readability}}.

\begin{figure}
    \centering
    \includegraphics[width=\columnwidth]{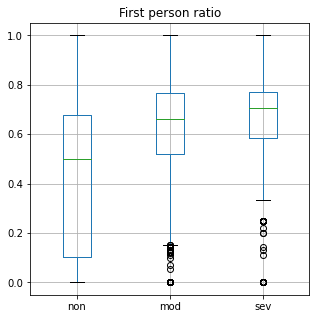}
    \caption{Usage ratio of the first person in posts with no depression signs, moderate signs and severe signs.}
    \label{fig:firstpratio}
\end{figure}

\paragraph{Person and number.} In addition, following previous research on the topic, we also decided to include information about the usage of first person, second person or third person and also singular vs. plural word distribution. The difference of the usage ratio of the first person is visualized in Figure \ref{fig:firstpratio} for posts with different levels of depression signs. In order to calculate those, we used the \texttt{stanza} library \cite{qi-etal-2020-stanza}.

\begin{center}
Example: \textit{\textbf{I} \textbf{am} lost because \textbf{I} do not like \textbf{them}}.
\end{center}

In this example there are three words that express information in first person, there is one word that is in third person and there is no word expressing the second person. Therefore, the vector encoding this information would be $(0.75, 0.0, 0.25)$. With regards to number, it finds that there are three singular form words and one expressing a plural form, thus the vector that encodes number will be $(0.75, 0.25)$. The final vector representing person and number is a concatenation of the previous two vectors ($[0.75, 0.0, 0.25, 0.75, 0.25]$).

%probability distr. of singular vs. plural, and 1st-2nd-3rd person.

\subsection*{Models}
As our goal was not to test how well different models would perform for the task, we decided to keep it simple and train Logistic Regression models. The main reason for doing this is the interpretability of the model, as the Logistic Regression is a relatively simple model.

We trained two different Logistic Regression models\footnote{All parameters are set to the default values.}\footnote{\url{https://scikit-learn.org/0.24/modules/generated/sklearn.linear_model.LogisticRegression.html}} with the following feature configuration:

\begin{itemize}
    \item Model 1: Words, POS-tags, Readability and style
    \item Model 2: Words, POS-tags, Readability and style, Person and number
\end{itemize}

\section{Results and Discussion}
Our best model, the second one, obtained a macro F1-score of $0.4429$ on the test data. The first model performed marginally worse with a macro F1-score of $0.439$. When performing our own experiments based on the training data, using a balanced train/test split, we had observed a rather higher performance, from which we could say that our model does not generalize well enough.

\begin{figure*}
    \centering
    \includegraphics[width=\textwidth,height=0.18\textheight]{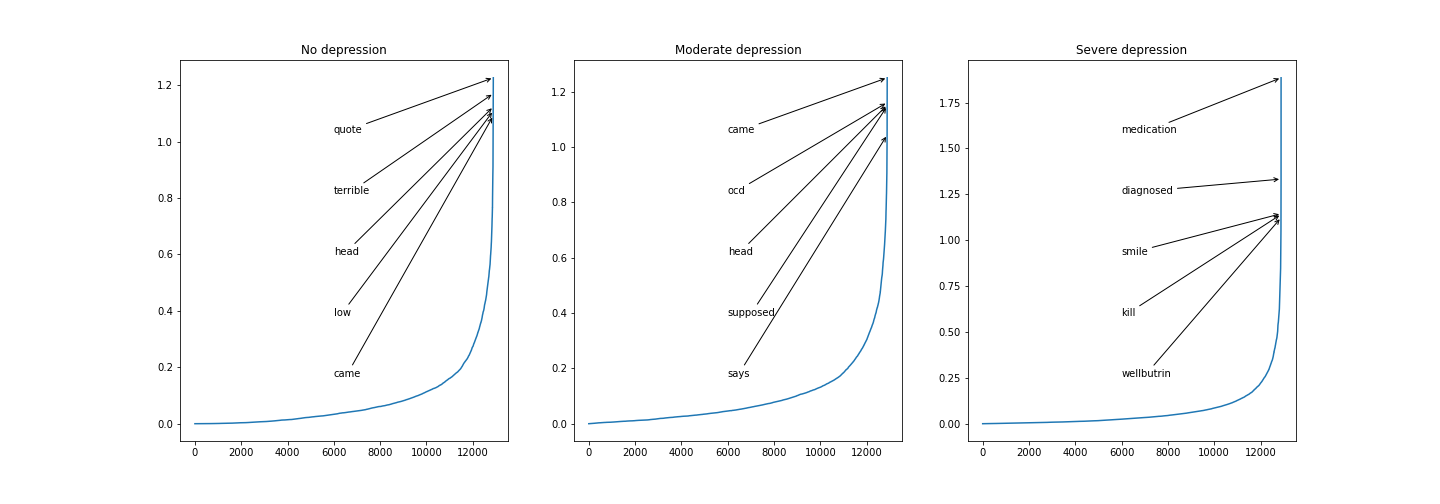}
    \caption{Sorted absolute values of Logistic Regression coefficients. We mark the rank of the top 5 features regarding words.}
    \label{fig:lrcoefswords}
\end{figure*}

\begin{figure*}
    \centering
    \includegraphics[width=\textwidth,height=0.18\textheight]{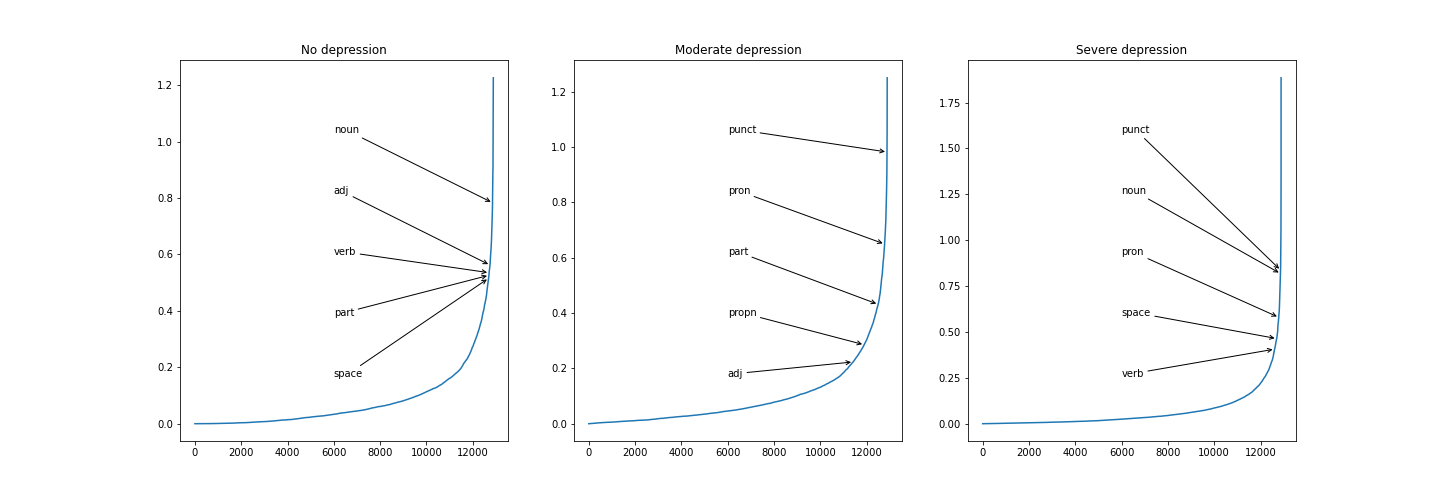}
    \caption{Sorted absolute values of Logistic Regression coefficients. We mark the rank of the top 5 features regarding POS tags.}
    \label{fig:lrcoefspos}
\end{figure*}

\begin{figure*}
    \centering
    \includegraphics[width=\textwidth,height=0.18\textheight]{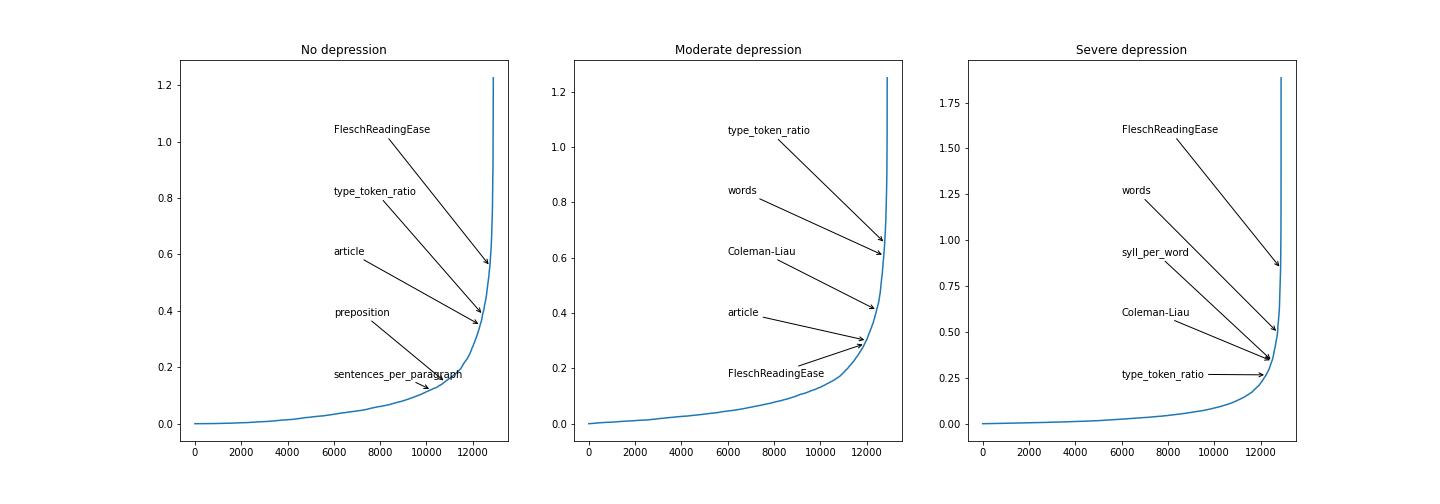}
    \caption{Sorted absolute values of Logistic Regression coefficients. We mark the rank of the top 5 features regarding readability \& style.}
    \label{fig:lrcoefsread}
\end{figure*}

\begin{figure*}
    \centering
    \includegraphics[width=\textwidth,height=0.18\textheight]{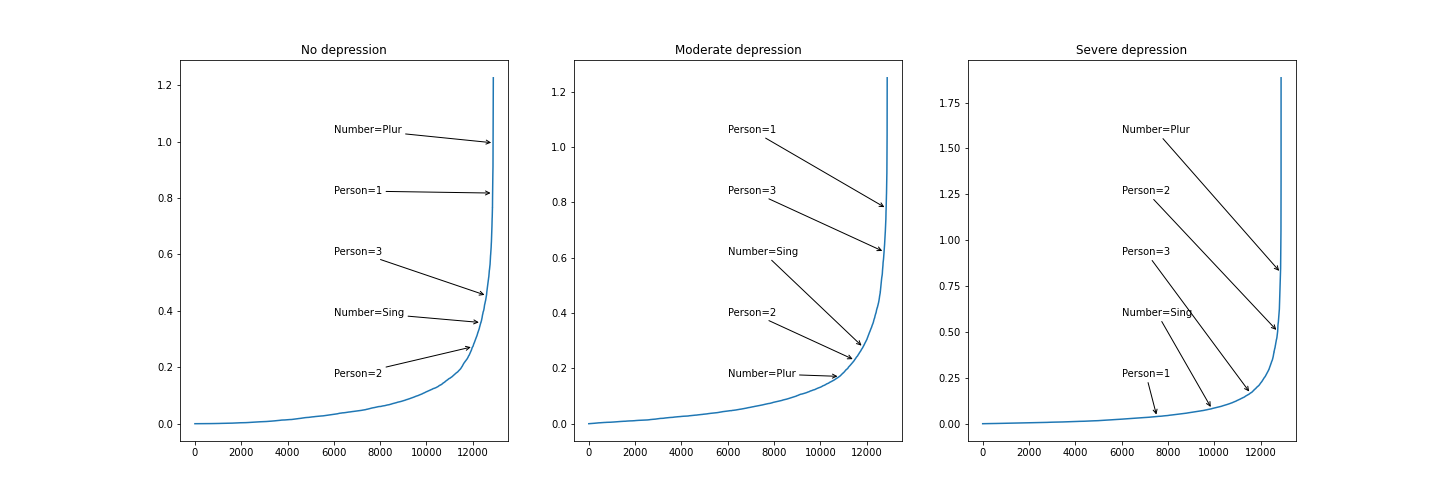}
    \caption{Sorted absolute values of Logistic Regression coefficients. We mark the rank of the top 5 features regarding person \& number.}
    \label{fig:lrcoefsnum}
\end{figure*}

From the results, and by comparing to the rest of participants, we can say that our model has several aspects to be improved. In the team wise classification our model ranked $25^{th}$, out of $31$ teams.

As the logistic regression model features are interpretable, we decided to analyze them more thoroughly, with the hope that this analysis is helpful for further research. For this analysis, we used the second model that makes use of the all the features and they were obtained after training the model with all available training data. Figures \ref{fig:lrcoefswords}, \ref{fig:lrcoefspos}, \ref{fig:lrcoefsread} and \ref{fig:lrcoefsnum} show the same sorted ranking of the features. In each figure we mark the position of the top 5 features, for each output class and for each feature template.\footnote{Our feature templates are words, POS-tags, readability \& style and person \& number.}

Figure \ref{fig:lrcoefspos} shows that punctuation marks and nouns are can be good predictors. In figure \ref{fig:lrcoefsread} we can observe that the \textit{Flesch Reading Ease} metric seems to be a good predictor together with the type token ratio. From figure \ref{fig:lrcoefsnum} we can observe that the first person ratio and the plural ratio seems to have a rather high effect in at least two classes of posts, meaning that they could be good predictors. Finally, figure \ref{fig:lrcoefswords} shows the importance of the top 5 words. These last features seem to have more importance than other features. This is because the vectorizer for words\footnote{We used \texttt{CountVectorizer} from Scikit-Learn.} was used in the default configuration and no normalization was done afterwards (all other features had values between $0$ and $1$). This means that at the current stage we cannot compare the importance of specific features across feature templates based on the coefficients of the model.

All the observations regarding feature importance should be taken with a grain of salt. A better approach would be to use a bootstrapping approach, training several models from subsets of the training corpus and analyzing the weight importance among several of those models.

\section{Conclusion and Future Work}
In this paper we presented our attempt to classify whether a social media post from Reddit shows signs of depression. We employed simple features and a linear model and we made an attempt to interpret the learned coefficients. As mentioned above, the model has several aspects that could be improved given its performance. Below we outline some possibilities for further research.

Following recent advances in Natural Language Processing, we think that including a pretrained word embedding model, such as BERT \cite{devlin-etal-2019-bert} would have positively contributed to the performance. These features could be additional features to the ones that we currently use or we could even fine-tune a pretrained model for this specific task.

Another aspect we believe that could improve the model is to include further syntactic information. The use of dependency parsing is being currently tested, but besides, there is also an extension of the \texttt{readability} package\footnote{\url{https://gist.github.com/andreasvc/1fcdcbc2a21d31722facd98e5f02d19a/}}, where syntactic information is obtained.

In addition to that, we expect that including the average sentiment of a post could be a relevant feature. Furthermore, recent advances in structured sentiment analysis\footnote{\url{https://competitions.codalab.org/competitions/33556}}\footnote{\url{https://github.com/jerbarnes/semeval22_structured_sentiment}} \cite{barnes-etal-2022-semeval} could potentially reveal mood changes.

%\vfill\null

\bibliography{anthology,custom}
\bibliographystyle{acl_natbib}

%\appendix

%\section{Example Appendix}
%\label{sec:appendix}

%This is an appendix.

\end{document}